\documentclass[10pt,twocolumn,letterpaper]{article}

\usepackage{cvpr}              %

\usepackage{graphicx}
\usepackage{amsmath}
\usepackage{amssymb}
\usepackage{booktabs}
\usepackage{subcaption}

\usepackage[pagebackref,breaklinks,colorlinks]{hyperref}

\usepackage[capitalize]{cleveref}
\crefname{section}{Sec.}{Secs.}
\Crefname{section}{Section}{Sections}
\Crefname{table}{Table}{Tables}
\crefname{table}{Tab.}{Tabs.}

\def\cvprPaperID{783} %
\def\confName{CVPR}
\def\confYear{2023}

\begin{document}

\title{Gloss Attention for Gloss-free Sign Language Translation}

\author{
Aoxiong Yin\textsuperscript{\rm 1} \thanks{Both authors contributed equally to this research.} ,
Tianyun Zhong\textsuperscript{\rm 1 *} ,
Li Tang\textsuperscript{\rm 1} ,
Weike Jin\textsuperscript{\rm 1} ,
Tao Jin\textsuperscript{\rm 1} ,
Zhou Zhao\textsuperscript{\rm 1}\thanks{Corresponding author.}\\
\small\textsuperscript{\rm 1}Zhejiang University\\
\tt\small \{yinaoxiong,zhongtianyun,tanglzju,weikejin,jint\_zju,zhaozhou\}@zju.edu.cn
}
\maketitle

\begin{abstract}
Most sign language translation (SLT) methods to date require the use of gloss annotations to provide additional supervision information, however, the acquisition of gloss is not easy.
To solve this problem, we first perform an analysis of existing models to confirm how gloss annotations make SLT easier.
We find that it can provide two aspects of information for the model, 1) it can help the model implicitly learn the location of semantic boundaries in continuous sign language videos, 2) it can help the model understand the sign language video globally.
We then propose \emph{gloss attention}, which enables the model to keep its attention within video segments that have the same semantics locally, just as gloss helps existing models do.
Furthermore, we transfer the knowledge of sentence-to-sentence similarity from the natural language model to our gloss attention SLT network (GASLT) to help it understand sign language videos at the sentence level.
Experimental results on multiple large-scale sign language datasets show that our proposed GASLT model significantly outperforms existing methods.
Our code is provided in \url{https://github.com/YinAoXiong/GASLT}.
\end{abstract}

\begin{figure}[t]
  \centering
  \subcaptionbox{self-attention\\+gloss-supervised\label{fig:joint_slt_att_v}}{\includegraphics[width = 0.3\linewidth]{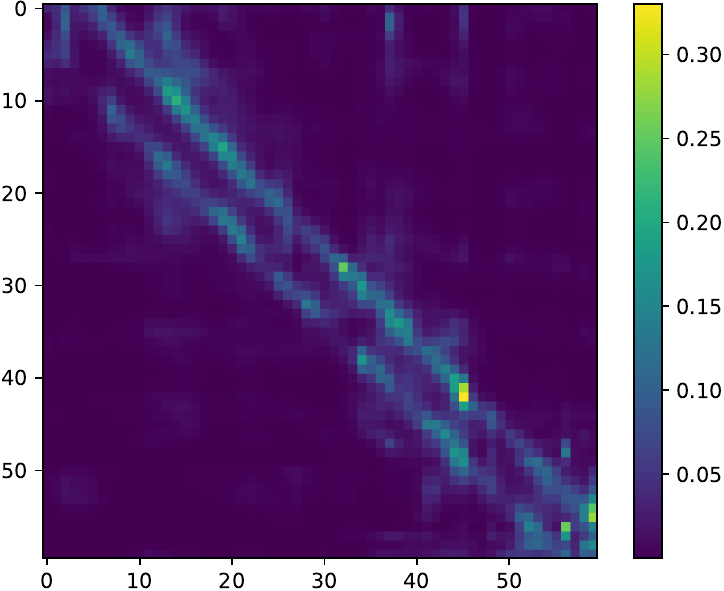}}
  \hfill
  \subcaptionbox{self-attention\\+gloss-free\label{fig:joint_slt_no_gloss}}{\includegraphics[width = 0.3\linewidth]{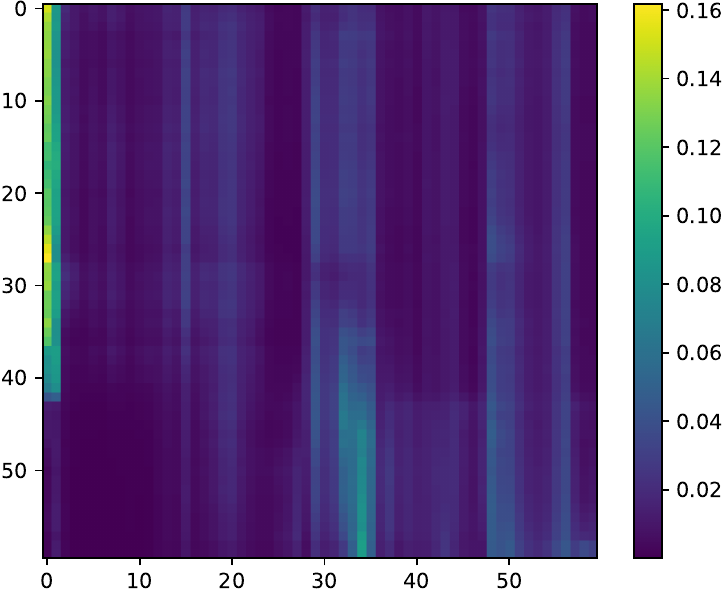}}
  \hfill
  \subcaptionbox{gloss-attention\\+gloss-free\label{fig:gaslt_no_gloss}}{\includegraphics[width = 0.3\linewidth]{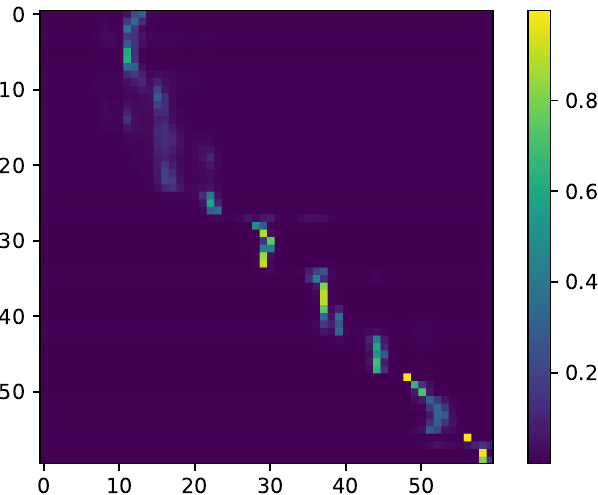}}
  \vspace{-2mm}
  \caption{Visualization of the attention map in the shallow encoder layer of three different SLT models. As shown in (a), an essential role of gloss is to provide alignment information for the model so that it can focus on relatively more important local areas. As shown in (b), the traditional attention calculation method is difficult to converge to the correct position after losing the supervision signal of the gloss. However, our proposed method (c) can still flexibly maintain the attention in important regions (just like (a)) due to the injection of inductive bias, which can partially replace the role played by gloss.}
  \label{fig:att_v}
\end{figure}

\section{Introduction}
\label{sec:intro}
Sign languages are the primary means of communication for an estimated 466 million deaf and hard-of-hearing people worldwide\cite{web1}.
Sign language translation (SLT), a socially important technology, aims to convert sign language videos into natural language sentences, making it easier for deaf and hard-of-hearing people to communicate with hearing people.
However, the grammatical differences between sign language and natural language \cite{pfau2018syntax,camgozNeuralSignLanguage2018} and the unclear semantic boundaries in sign language videos make it difficult to establish a mapping relationship between these two kinds of sequences.

Existing SLT methods can be divided into three categories, 1) two-stage gloss-supervised methods, 2) end-to-end gloss-supervised methods, and 3) end-to-end gloss-free methods.
The first two approaches rely on gloss annotations, chronologically labeled sign language words, to assist the model in learning alignment and semantic information.
However, the acquisition of gloss is expensive and cumbersome, as its labeling takes a lot of time for sign language experts to complete \cite{camgozNeuralSignLanguage2018}.
Therefore, more and more researchers have recently started to turn their attention to the end-to-end gloss-free approach \cite{liTSPNetHierarchicalFeature2020,orbayNeuralSignLanguage2020}.
It learns directly to translate sign language videos into natural language sentences without the assistance of glosses, which makes the approach more general while making it possible to utilize a broader range of sign language resources.
The gloss attention SLT network (GASLT) proposed in this paper is a gloss-free SLT method, which improves the performance of the model and removes the dependence of the model on gloss supervision by injecting inductive bias into the model and transferring knowledge from a powerful natural language model.

A sign language video corresponding to a natural language sentence usually consists of many video clips with complete independent semantics, corresponding one-to-one with gloss annotations in the semantic and temporal order.
Gloss can provide two aspects of information for the model. On the one hand, it can implicitly help the model learn the location of semantic boundaries in continuous sign language videos. 
On the other hand, it can help the model understand the sign language video globally.

In this paper, the GASLT model we designed obtain information on these two aspects from other channels to achieve the effect of replacing gloss.
First, we observe that the semantics of sign language videos are temporally localized, which means that adjacent frames have a high probability of belonging to the same semantic unit. 
The visualization results in Figure \ref{fig:joint_slt_att_v} and the quantitative analysis results in Table \ref{tab:diago} support this view.
Inspired by this, we design a new dynamic attention mechanism called gloss attention to inject inductive bias \cite{mitchell1980need} into the model so that it tends to pay attention to the content in the local same semantic unit rather than others.
Specifically, we first limit the number of frames that each frame can pay attention to, and set its initial attention frame to frames around it so that the model can be biased to focus on locally closer frames.
However, the attention mechanism designed in this way is static and not flexible enough to handle the information at the semantic boundary well.
We then calculate an offset for each attention position according to the input query so that the position of the model's attention can be dynamically adjusted on the original basis.
It can be seen that, as shown in Figure \ref{fig:gaslt_no_gloss}, our model can still focus on the really important places like Figure \ref{fig:joint_slt_att_v} after losing the assistance of gloss. In contrast, as shown in Figure \ref{fig:joint_slt_no_gloss}, the original method fails to converge to the correct position after losing the supervision signal provided by the gloss.

Second, to enable the model to understand the semantics of sign language videos at the sentence level and disambiguate local sign language segments, we transfer knowledge from language models trained with rich natural language resources to our model.
Considering that there is a one-to-one semantic correspondence between natural language sentences and sign language videos. We can indirectly obtain the similarity relationships between sign language videos by inputting natural language sentences into language models such as sentence bert \cite{reimersSentenceBERTSentenceEmbeddings2019}.
Using this similarity knowledge, we can enable the model to understand the semantics of sign language videos as a whole, which can partially replace the second aspect of the information provided by gloss.
Experimental results on three datasets RWTH-PHOENIX-WEATHER-2014T (PHOENIX14T)\cite{camgozNeuralSignLanguage2018}, CSL-Daily \cite{zhouImprovingSignLanguage2021} and SP-10 \cite{yinMLSLTMultilingualSign2022b} show that the translation performance of the GASLT model exceeds the existing state of the art methods, which proves the effectiveness of our proposed method.
We also conduct quantitative analysis and ablation experiments to verify the accuracy of our proposed ideas and the effectiveness of our model approach.

To summarize, the contributions of this work are as follows:
\begin{itemize}
  \item We analyze the role of gloss annotations in sign language translation.
  \item We design a novel attention mechanism and knowledge transfer method to replace the role of gloss in sign language translation partially.
  \item Extensive experiments on three datasets show the effectiveness of our proposed method. A broad range of new baseline results can guide future research in this field.
\end{itemize}

\section{Related Work}

\textbf{Sign Language Recognition.}
Early sign language recognition (SLR) was performed as isolated SLR, which aimed to recognize a single gesture from a cropped video clip \cite{huangSignLanguageRecognition2015, liWordlevelDeepSign2020, martinezPurdueRVLSLLLASL2002, ong2005automatic, starnerRealtimeAmericanSign1998, vaezijoze2019ms-asl,liTransferringCrossDomainKnowledge2020}.
Researchers then turned their interest to continuous SLR \cite{chengFullyConvolutionalNetworks2020, cihancamgozSubUNetsEndToEndHand2017, cuiDeepNeuralFramework2019, cuiRecurrentConvolutionalNeural2017, huangVideoBasedSignLanguage2018, minVisualAlignmentConstraint2021a, zhouSpatialTemporalMultiCueNetwork2020}, because this is the way signers actually use sign language.

\textbf{Sign Language Translation.}
The goal of SLT is to convert a sign language video into a corresponding natural language sentence \cite{camgozMultichannelTransformersMultiarticulatory2020, camgozNeuralSignLanguage2018, camgozSignLanguageTransformers2020, ganSkeletonAwareNeuralSign2021, koNeuralSignLanguage2019, liTSPNetHierarchicalFeature2020, orbayNeuralSignLanguage2020, yinBetterSignLanguage2020, yinSimulSLTEndtoEndSimultaneous2021, zhouImprovingSignLanguage2021, zhouSpatialTemporalMultiCueNetwork2020,chenSimpleMultiModalityTransfer2022,chentwo,jin2022mc,jin2021contrastive,jin2022prior}. Most existing methods use an encoder-decoder architecture to deal with this sequence-to-sequence learning problem.
Due to the success of the Transformer network in many fields \cite{huang2022transpeech, huanggenerspeech, huang2022prodiff, li2023date, lin2021simullr, xia2022video}, Camg{\"{o}}z et al. \cite{camgozSignLanguageTransformers2020} apply it to SLT and design a joint training method to use the information provided by gloss annotations to reduce the learning difficulty.
Zhou et al. \cite{zhouImprovingSignLanguage2021} propose a data augmentation method based on sign language back-translation to increase the SLT data available for learning. It first generates gloss text from natural language text and then uses an estimated gloss to sign bank to generate the corresponding sign sequence.
Yin et al. \cite{yinSimulSLTEndtoEndSimultaneous2021} propose a simultaneous SLT method based on the wait-k strategy \cite{maSTACLSimultaneousTranslation2019}, and they used gloss to assist the model in finding semantic boundaries in sign language videos.
Besides, some works improve the performance of SLT by considering multiple cues in sign language expressions \cite{camgozMultichannelTransformersMultiarticulatory2020, zhouSpatialTemporalMultiCueNetwork2020}.

\textbf{Gloss-free Sign Language Translation.}
Gloss-free SLT aims to train the visual feature extractor and translation model without relying on gloss annotations.
\cite{liTSPNetHierarchicalFeature2020} first explores the use of hierarchical structures to learn better video representations to reduce reliance on gloss.
Orbay et al. \cite{orbayNeuralSignLanguage2020} utilize adversarial, multi-task and transfer learning to search for semi-supervised tokenization methods to reduce dependence on gloss annotations.
\cite{voskou2021stochastic} proposes a new Transformer layer to train the translation model without relying on gloss.
However, the pre-trained visual feature extractor used by \cite{voskou2021stochastic} comes from \cite{koller2019weakly}, which uses the gloss annotation in the dataset during training.
The gloss-related information is already implicit in the extracted visual representations, so \cite{voskou2021stochastic} does not belong to the gloss-free SLT method.

\begin{figure*}[t]
  \includegraphics[width=\linewidth]{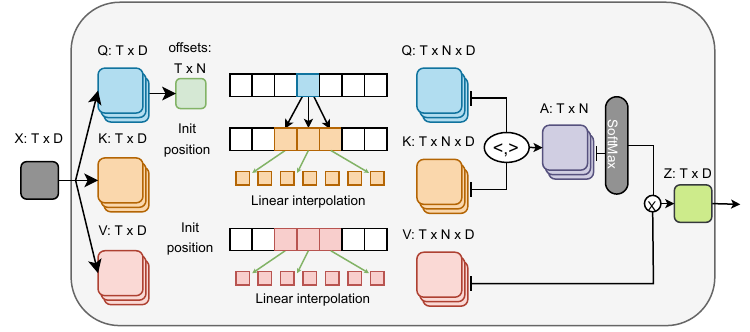}
  \caption{ \textbf{Gloss attention flowchart.} The initial focus of each query is the $N$ neighbors around it, and then the model will calculate $N$ offsets based on the query to adjust the focus position dynamically. Then use linear interpolation to get the final attention key and value. In this way, we make the model keep the attention in the correct position as it does with gloss supervision. The softmax operations are computed over the last dimension.}
  \label{fig:gloss_att}
\end{figure*}

\textbf{Sentence Embedding.}
Sentence embeddings aim to represent the overall meaning of sentences using dense vectors in a high-dimensional space \cite{aroraSimpleToughtobeatBaseline2017, conneauSupervisedLearningUniversal2017, kirosSkipThoughtVectors2015, leDistributedRepresentationsSentences2014, reimersSentenceBERTSentenceEmbeddings2019, ruckleConcatenatedPowerMean2018}.
Some early works use the linear combination of word embedding vectors in sentences to obtain sentence representations \cite{aroraSimpleToughtobeatBaseline2017, ruckleConcatenatedPowerMean2018}.
Subsequently, the emergence of large-scale self-supervised pre-trained language models such as BERT \cite{devlinBERTPretrainingDeep2019} significantly improves the effectiveness of natural language representation.
However, since BERT is not optimized for sentence embedding during pre-training, it does not perform well in sentence-level tasks such as text matching. The fact that BERT needs to input two sentences at the same time to calculate the similarity also makes the computational complexity high.
Sentence-BERT proposed by Reimers et al. \cite{reimersSentenceBERTSentenceEmbeddings2019} adopts the architecture of the Siamese network to solve this problem.
Since natural language has far more resources than sign language, in our work, we transfer knowledge from natural language models to sign language translation models.
This enables our model to understand sign language at the sentence level by learning the similarity between different sign language sentences.

\section{Analyzing The Role of Gloss in SLT}

In this section, we analyze and validate the idea we proposed in Section \ref{sec:intro} that gloss makes the attention map diagonal, and gloss helps the model understand the relationship between sign languages at the sentence level.

\begin{table}[htb]
  \caption{The degree of diagonalization of the attention map under different settings, the larger the CAD metrics, the higher the degree.}
  \label{tab:diago}
  \center
  \begin{tabular}{lccc}
    \toprule
    \textbf{Model} & \textbf{Layer1} & \textbf{Layer2} & \textbf{Layer3} \\
    \midrule
    gloss-supervised &0.9384&	0.7950&	0.7534 \\
    \midrule
    gloss-free & 0.8173 & 	0.7161&	0.6879    \\
    \bottomrule
    \end{tabular}
\end{table}

\paragraph{Quantitative Analysis of Diagonality.}
First inspired by \cite{shimUnderstandingRoleSelf2021}, we use \textit{cumulative attention diagonality} (CAD) metrics to quantitatively analyze the degree of diagonalization of attention maps in gloss-supervised and gloss-free settings.
As shown in Table \ref{tab:diago}, we can see that the degree of diagonalization of the attention map with gloss supervision is always higher than that of the attention map under the gloss-free setting.
This suggests that the attention map in the gloss-supervised setting is more diagonal, which is also what we observe when visualizing qualitative analysis, as shown in Figure \ref{fig:att_v}.

\begin{table}[htb]
  \caption{Average similarity difference metric for models under different settings.}
  \label{tab:asd}
  \vspace{-5mm}
  \center
  \begin{tabular}{ccc}
    \toprule
     & \textbf{gloss-supervised} & \textbf{gloss-free} \\
    \midrule
    $ASD$ &0.1593&	0.2815 \\
    \bottomrule
    \end{tabular}
\end{table}

\paragraph{Sign Language Sentence Embedding.}
We take the mean of the encoder output as the sign language sentence embedding and then use the cosine similarity to calculate the similarity of the two sentences.
We use the similarity between natural language sentences computed by sentence bert as the approximate ground truth. 
We evaluate whether gloss helps the model understand sign language videos at the sentence level by computing the \textit{average similarity difference} (ASD), that is, the difference between the similarity between the sign language sentence embedding and the natural language sentence embedding.
The calculation formula is as follows:

\begin{equation}
  ASD=\frac{1}{n^2-n}  \sum_{i = 1}^{n} \sum_{j=1}^{n} \left|  \widehat{S}[i,j] -S[i,j]  \right|  
\end{equation}
where $S[i,j]$ represents the similarity between natural language sentence embeddings, $\widehat{S}[i,j]$ represents the similarity between sign language sentence embeddings, and $n$ represents the number of sentence pairs.
As shown in Table \ref{tab:asd}, we can see that the ASD metric of the model is significantly lower than the model under the gloss-free setting when there is gloss supervision.
This shows that gloss annotations do help the model understand sign language videos at the sentence level.

\section{Methodology}
SLT is often considered a sequence-to-sequence learning problem \cite{camgozNeuralSignLanguage2018}.
Given a sign video $X' = (x'_1,x'_2,...,x'_T)$ with $T$ frames, SLT can be formulated as learning the conditional probability $p(Y'|X')$ of generating a spoken language sentence $Y' = (y'_1,y'_2,...,y'_M)$ with $M$ words.
We model translation from $X'$ to $Y'$ with Transformer architecture \cite{vaswaniAttentionAllYou2017}.
Our main contribution focuses on the encoder part, so we omit details about the decoder part, and the interested reader can refer to the original paper.
In this section, we first describe our designed gloss attention mechanism. Then we introduce how to transfer knowledge from natural language models to enhance the model's capture of global information in sign language videos.

\subsection{Embedding for Video and Text}
Similar to general sequence-to-sequence learning tasks, we first embed the input video and natural language text.
For the input video features, we follow a similar scheme as in \cite{camgozSignLanguageTransformers2020}. We simply use a linear layer to convert it to the dimension of the encoder, and then attach a relu \cite{glorotDeepSparseRectifier2011} activation function after batch normalization (BN) \cite{ioffeBatchNormalizationAccelerating2015} to get the embedded feature $\widehat{x}_t\in \mathbb{R}^D $.
For text embedding, we first use BPEmb \cite{DBLP:conf/lrec/Heinzerling018}, which is a BPE \cite{sennrichNeuralMachineTranslation2016} sub-word segmentation model learned on the Wikipedia dataset using the SentencePiece \cite{kudoSentencePieceSimpleLanguage2018} tool to segment text into sub-words.
BPE is a frequency-based sub-word division algorithm.
Dividing long words into subwords allows for generalized phonetic variants or compound words, which is also helpful for low-frequency word learning and alleviating out of vocabulary problems.
We then use the pre-trained sub-word embeddings in BPEmb as the initialization of the embedding layer and then convert the word vectors into text representations $\widehat{y}_m \in \mathbb{R}^D$ using a method similar to the visual feature embedding.
We formulate these operations as:
\begin{equation}
  \begin{split}
    \widehat{x}_t &= relu(BN(W_1x'_t+b_1)) + f_{pos}(t)\\
     \widehat{y}_m &= relu(BN(W_2Emb(y'_m)+b_2)) + f_{pos}(m)
  \end{split}
\end{equation}
Similar to other tasks, the position of a sign gesture in the whole video sequence is essential for understanding sign language.
Inspired by \cite{vaswaniAttentionAllYou2017,devlinBERTPretrainingDeep2019}, we inject positional information into input features using positional encoding $f_{pos}(\cdot )$.

\subsection{Gloss Attention}\label{sec:gloss_att}
After the operations in the previous section, we now have a set of tokens that form the input to a series of transformer encoder layers, as in the sign language transformer \cite{camgozSignLanguageTransformers2020}, consist of Layer Norm (LN) operations \cite{baLayerNormalization2016}, multi-head self-attention (MHSA) \cite{vaswaniAttentionAllYou2017}, residual connections \cite{heDeepResidualLearning2016}, and a feed-forward network (MLP):
\begin{equation}
  \begin{split}
    z &= MHSA(LN(x)) + x \\
    \widetilde{x} &= MLP(LN(z)) + z
  \end{split}
\end{equation}
Next, we discuss the difference between our proposed gloss attention and self-attention and how this inductive bias partially replaces the function of gloss.
For clarity, we use a single head in the attention operation as a demonstration in this section and ignore the layer norm operation.

For the self-attention operation, it first generates a set of $q_t, k_t, v_t \in \mathbb{R}^D $ vectors for each input sign language video feature $x_t$.
These vectors are computed as linear projections of the input $x_t$, that is, $q_t = W_qx_t$, $k_t = W_kx_t$, and $v_t = W_vx_t$, for each projection matrices $W_i \in \mathbb{R}^{D \times D} $.
Then the attention calculation result of each position is as follows:
\begin{equation}\label{e:self_att}
z_t
=
\sum_{i}^{T}
v_i \cdot
\frac
{\exp\langle q_t, k_i\rangle}
{\sum_{j}^{T}\exp \langle q_t, k_j\rangle}
\end{equation}
In this way, the attention score is calculated by dot products between each query $q_t$ and all keys $k_i$, and then the scores are normalized by softmax.
The final result is a weighted average of all the values $v_i$ using the calculated scores as weights.
Here for simplicity, we ignore the scaling factor $\sqrt{D} $ in the original paper and assume that all queries and keys have been divided by $\sqrt[4]{D} $.

There are two problems with this calculation. One is that its computational complexity is quadratic, as shown in Equation \ref{e:self_att}. The other more important problem is that its attention is difficult to converge to the correct position after losing the supervision of the gloss annotation, as shown in Figure \ref{fig:joint_slt_no_gloss}.
The root cause of this problem is that each query has to calculate the attention score with all keys. This approach can be very effective and flexible when strong supervision information is provided, but the model loses focus when supervision information is missing.

In order to solve the above problems, we propose \emph{gloss attention}, which is an attention mechanism we design according to the characteristics of sign language itself and the observation of the experimental results of existing models.
We observe that gloss-level semantics are temporally localized, that is, adjacent video frames are more likely to share the same semantics because they are likely to be in the same gloss-corresponding video segment.
Specifically, we first initialize $N$ attention positions $P = (p_1,p_2,...,p_N)$ for each qeury, where $p_1 = t-\left\lceil N/2\right\rceil $, $p_n=t+N-\left\lceil N/2\right\rceil $, and the intermediate interval is 1.
Later, in order to better deal with the semantic boundary problem, we will calculate the $N$ offset according to the input query to dynamically adjust the position of the attention:
\begin{equation}
  O = W_oq_t; \hspace{12pt} \widehat{P} = (P+O)\%T
\end{equation}
where $W_o \in \mathbb{R}^{N\times D} $, $\widehat{P}$ is the adjusted attention position, and we take the remainder of $T$ to ensure that the attention position will not cross the bounds.
The adjusted attention positions have become floating-point numbers due to the addition of offset $O$, and the indices of keys and values in the array are integers.
For this reason, we use linear interpolation to get the keys $\widehat{K_t} = (\widehat{k_t^1},\widehat{k_t^2},...,\widehat{k_t^N} ) $ and values $\widehat{V_t} = (\widehat{v_t^1},\widehat{v_t^2},...,\widehat{v_t^N} ) $ that are finally used for calculation:
\begin{equation}
  b_i = \left\lfloor \widehat{p_i} \right\rfloor, u_i = b_i+1 
\end{equation}
\begin{equation}
  \begin{split}
    \widehat{k_t^i} = (u_i-\widehat{p_i} )\cdot  k_{b_i} + (\widehat{p_i}-b_i)\cdot k_{u_i}\\  
    \widehat{v_t^i} = (u_i-\widehat{p_i} )\cdot  v_{b_i} + {p_i}-b_i)\cdot v_{u_i} 
  \end{split}
\end{equation}
Finally, the attention calculation method for each position is as follows:
\begin{equation}\label{e:gloss_att}
  z_t
  =
  \sum_{i}^{N}
  \widehat{v_t^i}  \cdot
  \frac
  {\exp\langle q_t, \widehat{k_t^i} \rangle}
  {\sum_{j}^{T}\exp \langle q_t, \widehat{k_t^j} \rangle}
\end{equation}
Compared with the original self-attention, the computational complexity of gloss attention is $\mathcal{O} (NT)$, where $N$ is a constant and in general $N \ll T$, so the computational complexity of gloss attention is $\mathcal{O} (n)$.
In addition, as shown in Figure \ref{fig:gaslt_no_gloss}, the visualization results show that the gloss attention we designed can achieve similar effects to those with gloss supervision. The experimental results in Section \ref{sec:exp} also demonstrate the effectiveness of our proposed method.
A flowchart of the full gloss attention operation is shown in tensor form in Figure \ref{fig:gloss_att}.

\begin{table*}[t]
  \caption{Comparisons of gloss-free translation results on RWTH-PHOENIX-Weather 2014T dataset. }
  \label{tab:gloss-free}
  \setlength{\tabcolsep}{4mm}
  \centering
  \begin{tabular}{lccccc}
    \toprule
    \textbf{Methods}  & \textbf{ROUGE-L} &\textbf{ BLEU-1} & \textbf{BLEU-2} & \textbf{BLEU-3} & \textbf{BLEU-4} \\
    \midrule
    Conv2d-RNN~ \cite{camgozNeuralSignLanguage2018}  & 29.70 &27.10 & 15.61 & 10.82 & 8.35 \\
    ~~~~~~+ Luong Attn.~\cite{camgozNeuralSignLanguage2018}+\cite{luongEffectiveApproachesAttentionbased2015}& 30.70 & 29.86 & 17.52 & 11.96 & 9.00 \\
    ~~~~~~+ Bahdanau Attn.~\cite{camgozNeuralSignLanguage2018}+\cite{bahdanauNeuralMachineTranslation2015} & 31.80 & 32.24 &  19.03 & 12.83 & 9.58 \\
    \midrule
    Joint-SLT~\cite{camgozSignLanguageTransformers2020} &31.10 & 30.88 & 18.57 & 13.12 & 10.19 \\
    \midrule
    Tokenization-SLT~\cite{orbayNeuralSignLanguage2020} &36.28 & 37.22 & 23.88 & 17.08 & 13.25 \\
    \midrule
    TSPNet-Sequential~\cite{liTSPNetHierarchicalFeature2020}  &34.77 & 35.65 & 22.80 & 16.60 & 12.97 \\
    TSPNet-Joint~\cite{liTSPNetHierarchicalFeature2020}  &34.96 & 36.10 & 23.12 & 16.88 & 13.41 \\
    \midrule
    \textbf{GASLT}  &\textbf{39.86} & \textbf{39.07} & \textbf{26.74} & \textbf{21.86} & \textbf{15.74} \\
    \bottomrule
  \end{tabular}
\end{table*}
\begin{table*}[t]
  \caption{Comparisons of gloss-free translation results on CSL-Daily (top) and SP-10 (bottom) datasets. }
  \label{tab:gloss-free-sp10-csl}
  \begin{subtable}{\linewidth}
    \centering
    \setlength{\tabcolsep}{5mm}
    \begin{tabular}{lccccc}
      \toprule
      \textbf{Methods}  & \textbf{ROUGE-L} &\textbf{ BLEU-1} & \textbf{BLEU-2} & \textbf{BLEU-3} & \textbf{BLEU-4} \\
      \midrule
      Joint-SLT~\cite{camgozSignLanguageTransformers2020} &19.61 & 21.56 & 8.29 & 3.68 & 1.72 \\
      \midrule
      TSPNet-Joint~\cite{liTSPNetHierarchicalFeature2020}  &18.38 & 17.09 & 8.98 & 5.07 & 2.97 \\
      \midrule
      \textbf{GASLT}  &\textbf{20.35} & \textbf{19.90} & \textbf{9.94} & \textbf{5.98} & \textbf{4.07} \\
      \bottomrule
    \end{tabular}
  \end{subtable}
  \begin{subtable}{\linewidth}
    \centering
    \setlength{\tabcolsep}{5mm}
    \begin{tabular}{lccccc}
      \toprule
      \textbf{Methods}  & \textbf{ROUGE-L} &\textbf{ BLEU-1} & \textbf{BLEU-2} & \textbf{BLEU-3} & \textbf{BLEU-4} \\
      \midrule
      Joint-SLT~\cite{camgozSignLanguageTransformers2020} &12.23 & 12.49 & 6.79 & 3.99 & 1.60 \\
      \midrule
      TSPNet-Joint~\cite{liTSPNetHierarchicalFeature2020}  &15.18 & 13.55 & 7.07 & 3.77 & 2.20 \\
      \midrule
      \textbf{GASLT}  &\textbf{16.98} & \textbf{21.72} & \textbf{10.92} & \textbf{6.61} & \textbf{4.35} \\
      \bottomrule
    \end{tabular}
  \end{subtable}
\end{table*}

\subsection{Knowledge Transfer}
\label{sec:kt}
Another important role of gloss is to help the model understand the entire sign language video from a global perspective. Its absence will reduce the model's ability to capture global information.
Fortunately, however, we have language models learned on a rich corpus of natural languages, and they have been shown to work well on numerous downstream tasks. 
Since there is a one-to-one semantic relationship between sign language video and annotated natural language text, we can transfer the knowledge from the language model to our model.
Specifically, we first use sentence bert \cite{reimersSentenceBERTSentenceEmbeddings2019} to calculate the cosine similarity $ S \in \mathbb{R}^{D_t\times D_t} $ between all natural language sentences offline, where $D_t$ is the size of the training set.
Then we aggregate all the video features output by the encoder to obtain an embedding vector $e \in \mathbb{R}^D$ representing the entire sign language video.
There are various ways to obtain the embedding vector, and we analyze the impact of choosing different ways in Section \ref{sec:ab}.
Finally we achieve knowledge transfer by minimizing the mean squared error of cosine similarity between video vectors and cosine similarity between natural languages:
\begin{equation}
  \mathcal{L}_{kt} =\left(\frac{e_i\cdot e_j}{\|e_i\|\|e_j\| } - S[i,j] \right)^2 
\end{equation}
In this way we at least let the model know which sign language videos are linguistically similar and which are semantically different.

\section{Experiments} \label{sec:exp}

\subsection{Experiment Setup and Implementation Details}\label{sec:ex_details}

\textbf{Datasets.}
We evaluate the GASLT model on the RWTH-PHOENIX-WEATHER-2014T (PHOENIX14T)\cite{camgozNeuralSignLanguage2018}, CSL-Daily \cite{zhouImprovingSignLanguage2021} and SP-10 \cite{yinMLSLTMultilingualSign2022b} datasets.
We mainly conduct ablation studies and experimental analysis on the PHOENIX14T dataset.
PHOENIX14T contains weather forecast sign language videos collected from the German public television station PHOENIX and corresponding gloss annotations and natural language text annotations to these videos.
CSL-Daily is a recently released large-scale Chinese sign language dataset, which mainly contains sign language videos related to daily life, such as travel, shopping, medical care, etc.
SP-10 is a multilingual sign language dataset that contains sign language videos in 10 languages.
For all datasets we follow the official partitioning protocol.

\textbf{Evaluation Metrics.}
Similar to previous papers, we evaluate the translation performance of our model using BLEU \cite{papineniBleuMethodAutomatic2002} and ROUGE-L \cite{linAutomaticEvaluationMachine2004} scores, two of the most commonly used metrics in machine translation.
BLEU-n represents the weighted average translation precision up to n-grams. Generally, we use uniform weights, that is, the weights from 1-grams to n-grams are all $1/n$.
ROUGE-L uses the longest common subsequence between predicted and reference texts to calculate the F1 score.
We use the script officially provided by the Microsoft COCO caption task \cite{chenMicrosoftCOCOCaptions2015} to calculate the ROUGE-L score, which sets $\beta = 1.2$ in the F1 score
\footnote{\url{https://github.com/tylin/coco-caption}}.

\textbf{Implementation and Optimization.}
We use the pytorch \cite{paszkePyTorchImperativeStyle2019} framework to implement our GASLT model based on the open source code of \cite{kreutzer-etal-2019-joey} and \cite{camgozSignLanguageTransformers2020}.
Our model is based on the Transformer architecture, the number of hidden units in the model, the number of heads, and the layers of encoder and decoder are set to 512, 8, 2, 2, respectively.
The parameter $N$ in gloss attention is set to 7.
We also use dropout with 0.5 and 0.5 drop rates on encoder and decoder layers to mitigate overfitting.
For a fair comparison, we uniformly use the pre-trained I3D model in TSPNet \cite{liTSPNetHierarchicalFeature2020} to extract visual features.
For models other than TSPNet, we only use visual features extracted with a sliding window of eight and stride of two.
We adopt Xavier initialization \cite{glorotUnderstandingDifficultyTraining2010} to initialize our network.
we use label smoothed \cite{szegedyRethinkingInceptionArchitecture2016} crossentropy loss to optimize the SLT task, where the smoothing parameter $\varepsilon $ is set to $0.4$.
We set the batch size to 32 when training the model.
We use the Adam \cite{kingmaAdamMethodStochastic2015} optimizer with an initial learning rate of $5\times10^{-4}$ ($\beta_1$=0.9, $\beta_2$=0.998, $\epsilon =10^{-8} $), and the weight decays to $10^{-3}$.
We use similar plateau learning rate scheduling as in \cite{camgozSignLanguageTransformers2020}, except we adjust the patience and decrease factor to 9 and 0.5, respectively. 
The weights of translation cross-entropy loss and knowledge transfer loss $\mathcal{L}_{kt}$ are both set to one.
All experiments use the same random seed.

\subsection{Comparisons with the State-of-the-art}

\textbf{Competing Methods.}
We compare our GASLT model with three gloss-free SLT methods. 
1) Conv2d-RNN \cite{camgozNeuralSignLanguage2018} is the first proposed gloss-free SLT model, which uses a GRU-based \cite{choLearningPhraseRepresentations2014} encoder-decoder architecture for sequence modeling.
2) Tokenization-SLT \cite{orbayNeuralSignLanguage2020} achieves the state-of-the-art on the ROUGE score of PHOENIX14T dataset, which utilizes adversarial, multi-task, and transfer learning to search for semi-supervised tokenization methods to reduce dependence on gloss annotations.
3) Joint-SLT \cite{camgozSignLanguageTransformers2020} is the first sign language translation model based on the Transformer architecture, which jointly learns the tasks of sign language recognition and sign language translation.
4) TSPNet \cite{liTSPNetHierarchicalFeature2020} achieves the state-of-the-art on the BLEU score of PHOENIX14T dataset, which enhances translation performance by learning hierarchical features in sign language.

\definecolor{navyblue}{rgb}{0.17, 0.22, 0.71}
\definecolor{lava}{rgb}{0.81, 0.06, 0.13}
\begin{table}[b]
    \caption{Comparison of the example gloss-free translation results of GASLT and the previous state-of-the-art model. We highlight correctly translated 1-grams in \textcolor{navyblue}{{blue}}, semantically correct translation in \textcolor{lava}{{red}}.}
    \label{tab:exp}
    \centering
    \resizebox{\linewidth}{!}{
    \begin{tabular}{rll}
        
  \toprule   
     Ground Truth: & und nun die wettervorhersage für morgen donnerstag den siebzehnten dezember . \\
     & ( and now the weather forecast for tomorrow thursday the seventeenth of december . ) \\
     TSPNet \cite{liTSPNetHierarchicalFeature2020}: & \textcolor{navyblue}{{und nun die wettervorhersage für morgen donnerstag den}} sechzehnten januar . \\
     & ( \textcolor{navyblue}{{and now the weather forecast for tomorrow thursday the}} sixteenth of january . ) \\
     Ours: & \textcolor{navyblue}{{und nun die wettervorhersage für morgen donnerstag den siebzehnten dezember}} . \\
     & ( \textcolor{navyblue}{{and now the weather forecast for tomorrow thursday the seventeenth of december}} . ) \\
     \midrule
     
     Ground Truth: & es gelten entsprechende warnungen des deutschen wetterdienstes .\\
     & ( Appropriate warnings from the German Weather Service apply . ) \\
     TSPNet \cite{liTSPNetHierarchicalFeature2020}: & am montag gibt \textcolor{navyblue}{es} hier und da schauer in der südwesthälfte viel sonne . \\
     & ( on monday there will be showers here and there in the south-west half, lots of sun . ) \\
     Ours: & \textcolor{navyblue}{{es gelten entsprechende }} \textcolor{lava}{unwetterwarnungen} \textcolor{navyblue}{des deutschen wetterdienstes .} \\
     & ( \textcolor{navyblue}{Appropriate} \textcolor{lava}{severe weather warnings} \textcolor{navyblue}{from the German Weather Service apply .} )\\
     \midrule
     Ground Truth: & morgen reichen die temperaturen von einem grad im vogtland bis neun grad am oberrhein .\\
     & ( tomorrow the temperatures will range from one degree in the vogtland to nine degrees on the upper rhine . ) \\
     TSPNet \cite{liTSPNetHierarchicalFeature2020}: & heute nacht zehn grad am oberrhein und fünf \textcolor{navyblue}{grad am oberrhein} . \\
     & ( tonight ten \textcolor{navyblue}{degrees on the upper rhine} and five degrees on the upper rhine. ) \\
     Ours: & \textcolor{navyblue}{morgen temperaturen von} null \textcolor{navyblue}{grad im vogtland bis neun grad am oberrhein .} \\
     & ( \textcolor{navyblue}{tomorrow temperatures from} zero \textcolor{navyblue}{degrees in the vogtland to nine degrees on the upper rhine} . )\\
     \bottomrule
    \end{tabular}
    }
\end{table}

\textbf{Quantitative Comparison.}
We report the BLEU scores and ROUGE scores of our GASLT model and comparison models on the PHOENIX14T dataset in Table \ref{tab:gloss-free}.
For Joint-SLT we reproduce and report its results in the gloss-free setting; for other models, we use the data reported in the original paper.
As shown in Table \ref{tab:gloss-free}, the translation performance of our model significantly outperforms the original two state-of-the-art gloss-free SLT models, Tokenization-SLT and TSPNet-Joint, the blue4 score is improved from 13.41 to 15.74 (17.37\%), and the  ROUGE-L score is improved from 36.28 to 39.86 (9.86\%).
As shown in Table \ref{tab:gloss-free-sp10-csl}, we further evaluate our proposed GASLT model on two other public datasets, and we can see that our method outperforms existing methods on both datasets.
Benefiting from the injection of prior information about semantic temporal locality in our proposed gloss attention mechanism and its flexible attention span, our GASLT model can keep attention in the right place.
Coupled with the help of knowledge transfer, the GASLT model significantly narrows the gap between gloss-free SLT and gloss-supervised SLT methods compared to previous gloss-free SLT methods.

\textbf{Qualitative Comparison.}
We present 3 example translation results generated by our GASLT model and TSPNet model in Table \ref{tab:exp} for qualitative analysis.
In the first example, our model produces a very accurate translation result, while TSPNet gets the date wrong.
In the second example, our model ensures that the semantics of the sentence has not changed by using the synonym of "warnungen" (warnings) such as "unwetterwarnungen" (severe weather warnings), while TSPNet has a translation error and cannot correctly express the meaning of the sign language video.
In the last example, it can be seen that although our generated results differ in word order from the ground truth, they express similar meanings.
However, existing evaluation metrics can only make relatively mechanical comparisons, making it difficult to capture these differences.
We provide the full translation results generated by our proposed model in the supplementary material.

\begin{table}[htb]
  \centering
  \setlength{\abovecaptionskip}{2pt}
  \caption{Results of ablation experiments on the PHOENIX14T dataset. KT represents the knowledge transfer method proposed in Section \ref{sec:kt}.}
  \label{tab:ab}
  \resizebox{\linewidth}{!}{
  \begin{tabular}{lccccc}
    \toprule  %
    \textbf{Model}&\textbf{R}&\textbf{B1}&\textbf{B2}&\textbf{B3}&\textbf{B4}\\
    \midrule  %
    self-attention              &28.53&30.05&18.08&12.71&9.78\\
    +KT           &36.53&35.86&23.09&16.46&12.66\\
    \midrule
    sliding window attention \cite{beltagy2020longformer}             &33.48&31.83&20.31&14.68&11.46\\
    +KT          &38.46&37.67&24.82&18.06&14.07\\
    \midrule
    dilated sliding window attention\cite{beltagy2020longformer}            &33.80&30.08&19.16&13.84&10.82\\
    +KT          &38.16&34.58&23.17&17.29&13.78\\
    \midrule
    global+sliding window attention\cite{beltagy2020longformer}           &36.73&33.05&21.39&15.63&12.22\\
    +KT          &38.86&36.37&24.34&17.98&14.17\\
    \midrule
    BIGBIRD attention  \cite{zaheer2020big}         &36.19&33.08&21.59&15.69&12.33\\
    +KT          &38.67&35.69&23.92&17.77&14.06\\
    \midrule
    Gloss attention             &38.24&37.26&25.18&18.80&14.93\\
    +KT (GASLT) &\textbf{39.86} & \textbf{39.07} & \textbf{26.74} & \textbf{21.86} & \textbf{15.74} \\
    \bottomrule %
    \end{tabular}
  }
    \label{table:ab}
\end{table}
\vspace{-5mm}
\subsection{Ablation studies}\label{sec:ab}
In this section, we introduce the results of our ablation experiments on the PHOENIX14T dataset, and analyze the effectiveness of our proposed method through the experimental results.
In addition, we also study the impact of different component choices and different parameter settings on the model performance.
To facilitate the expression, in the table in this section, we use R to represent ROUGE-L, B1$\rightarrow$B4 to represent BLEU1$\rightarrow$BLEU4.  

\textbf{The Effectiveness of Gloss Attention.}
As shown in Table \ref{tab:ab}, we test the model's performance with self-attention, local-attention, and gloss-attention, respectively, on the PHOENIX14T dataset, where local-attention and gloss-attention use the same window size.
We can see that local-attention performs better than self-attention, while gloss-attention achieves better performance than both.
This shows that the attention mechanism of gloss-attention, which introduces inductive bias without losing flexibility, is more suitable for gloss-free sign language translation.

\textbf{The Effectiveness of Knowledge Transfer.}
As shown in Table \ref{tab:ab}, we add our proposed knowledge transfer method to various attention mechanisms, and we can see that it has an improved effect on all attention mechanisms. This demonstrates the effectiveness of our proposed knowledge transfer method.

\begin{table}[htb]
  \caption{Analyze the impact of the number of initialized attention positions $N$ in gloss attention on model performance. We report ROUGE-L scores in R column; BLEU-$n$ in B-$n$ columns.}
  \vspace{-5mm}
  \label{tab:ab_N_no_gloss}
  \center
  \resizebox{0.9\linewidth}{!}{
  \begin{tabular}{lcccccc}
    \toprule
    $N$& \textbf{R} &\textbf{ B1} & \textbf{B2} & \textbf{B3} & \textbf{B4} \\
    \midrule
      3 & 39.19 & 37.68 &25.49 & 19.03 & 15.16 \\
      \midrule
      5 & 39.62 & 38.34 &26.06 & 19.53 & 15.50 \\
      \midrule
      7 & \textbf{39.86} & \textbf{39.07} & \textbf{26.74} & \textbf{21.86} & \textbf{15.74} \\
      \midrule
      9 &39.41 & 38.24 &25.73 & 19.10 & 15.07 \\
      \midrule
      11 &39.14 & 38.25 &25.57 & 18.93 & 14.95 \\
    \bottomrule
    \end{tabular}
  }
\end{table}

\textbf{Gloss Attention.}
We then explore the effect of the number $N$ of initialized attention positions in gloss attention on model performance.
As shown in Table \ref{tab:ab_N_no_gloss}, without using gloss, the BLEU-4 score of the model increases first and then decreases with the increase of $N$, and reaches the best performance when $N=5$.
This demonstrates that too few attention positions will limit the expressive ability of the model, while too large $N$ may introduce interference information.
After all, when $N=T$, the calculation method of gloss attention will be no different from the original self-attention.
In addition, the translation performance of the model is the best when $N=7$ (due to the introduction of linear interpolation, the actual field of view of the model at this time is 14), which is also close to the statistics of 15 video frames per gloss in the PHOENIX14T dataset.

\begin{table}[h]
  \caption{Comparison between different sign language sentence embedding vector generation methods. }
  \label{tab:ab_sentence_emb}
  \vspace{-3mm}
  \centering
  \resizebox{\linewidth}{!}{
  \begin{tabular}{lccccc}
    \toprule
    \textbf{Methods}  & \textbf{R} &\textbf{ B1} & \textbf{B2} & \textbf{B3} & \textbf{B4} \\
    \midrule
    CLS-vector & 38.50 & 37.18 &24.99 & 18.59 & 14.70 \\ %
    \midrule
    Ave. gloss-attention\\ embedding & \textbf{39.86} & \textbf{39.07} & \textbf{26.74} & \textbf{21.86} & \textbf{15.74} \\
    \midrule
    Max. gloss-attention\\ embedding &35.63 & 35.78 &22.74 & 16.40 & 12.76 \\
    \midrule
    Ave. self-attention\\ embedding &37.58 & 37.68 &24.67 & 17.92 & 13.97 \\
    \midrule
    Max. self-attention\\ embedding &35.54 & 34.40 &21.85 & 15.89 & 12.29 \\
    \bottomrule
  \end{tabular}
  }
\end{table}
\textbf{Sign Language Sentence Embedding.}
Then we compare the impact of different sign language sentence embedding vector generation methods on the model performance.
The experimental results are shown in Table \ref{tab:ab_sentence_emb}.
In the table, CLS-vector indicates that a special CLS token is used to aggregate global information as the sentence embedding. Ave demonstrates that the average of all the vectors output by the encoder is used as the sentence embedding. Max means to take the maximum value of each dimension for all the vectors output by the encoder as the sentence embedding. Gloss attention embedding means that only gloss attention is used in the encoder. Self-attention embedding means that a layer using self-attention is added at the end of the encoder.
It can be seen that the sentence embedding generated by the method of CLS-vector does not perform well in the model performance.
In addition, we can find that the Ave method performs better in translation performance than the Max method.
The model achieves the best performance when using the Ave. gloss-attention embedding method, which demonstrates that thanks to the superposition of receptive fields and the flexible attention mechanism, the model can capture global information well even when only gloss attention is used.

\begin{figure}[t]
  \centering
  \includegraphics[width=\linewidth]{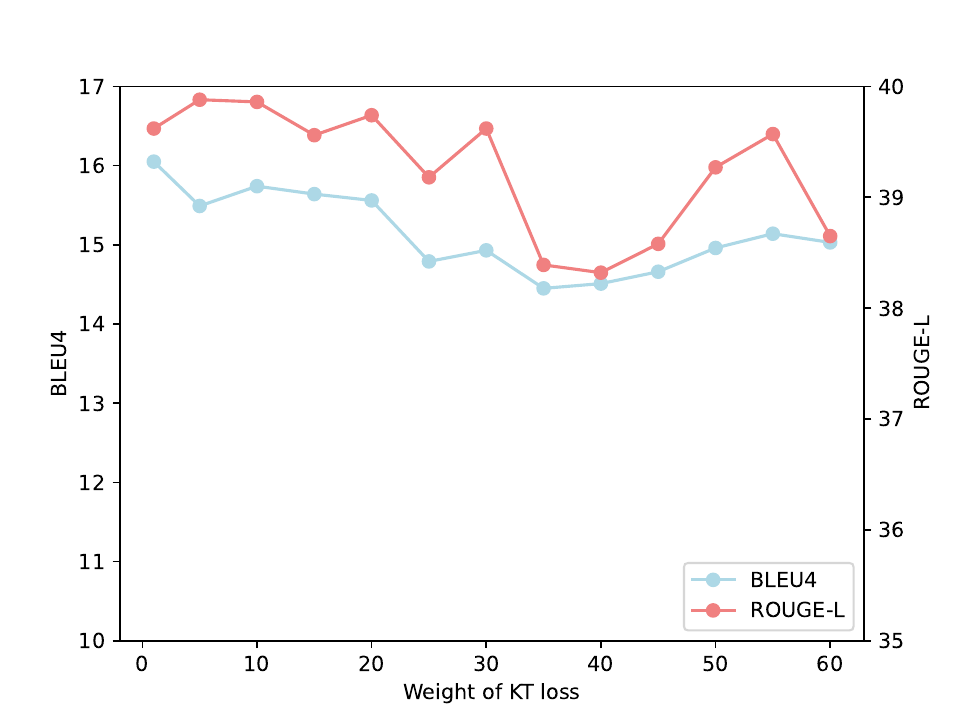}
  \vspace{-6mm}
  \caption{The curve of model performance with the weight of knowledge transfer loss.}
  \label{fig:weight_loss}
\end{figure}

\textbf{Weight of Knowledge Transfer Loss.}
Finally, we analyze the effect of setting different weights for the knowledge transfer loss on the model performance.
As shown in Figure \ref{fig:weight_loss}, we can find that the model's performance tends to decrease as the weight of the knowledge transfer loss increases.
This may be because the similarity relationship between sentences obtained from Sentence Bert is not so accurate, and too high weight will cause the model to overfit the similarity relationship and decrease translation performance.

\section{Conclusion}
In this paper, we analyze the role of gloss annotations in the SLT task.
Then we propose a new attention mechanism, gloss attention, which can partially replace the function of gloss. 
The gloss attention, which is designed according to the temporal locality principle of sign language semantics, enables the model to keep the attention within the video segments corresponding to the same semantics, just as the supervision signal provided by the gloss is still there.
In addition, we design a new knowledge transfer method to help the model better capture global sign language semantics.
In the appendix, we discuss the limitations of our work.

\section*{Acknowledgments}
This work was supported by National Natural Science Foundation of China under Grant No. 62222211, No.61836002 and No.62072397.

{\small
\bibliographystyle{ieee_fullname}
\bibliography{cvpr2023_references,manual}

\begin{thebibliography}{10}\itemsep=-1pt

\bibitem{aroraSimpleToughtobeatBaseline2017}
Sanjeev Arora, Yingyu Liang, and Tengyu Ma.
\newblock A simple but tough-to-beat baseline for sentence embeddings.
\newblock In {\em 5th International Conference on Learning Representations,
  {{ICLR}} 2017, Toulon, France, April 24-26, 2017, Conference Track
  Proceedings}. {OpenReview.net}, 2017.

\bibitem{baLayerNormalization2016}
Jimmy~Lei Ba, Jamie~Ryan Kiros, and Geoffrey~E. Hinton.
\newblock Layer {{Normalization}}, July 2016.

\bibitem{bahdanauNeuralMachineTranslation2015}
Dzmitry Bahdanau, Kyunghyun Cho, and Yoshua Bengio.
\newblock Neural machine translation by jointly learning to align and
  translate.
\newblock In Yoshua Bengio and Yann LeCun, editors, {\em 3rd International
  Conference on Learning Representations, {{ICLR}} 2015, San Diego, {{CA}},
  {{USA}}, May 7-9, 2015, Conference Track Proceedings}, 2015.

\bibitem{beltagy2020longformer}
Iz Beltagy, Matthew~E Peters, and Arman Cohan.
\newblock Longformer: The long-document transformer.
\newblock {\em arXiv preprint arXiv:2004.05150}, 2020.

\bibitem{camgozNeuralSignLanguage2018}
Necati~Cihan Camgoz, Simon Hadfield, Oscar Koller, Hermann Ney, and Richard
  Bowden.
\newblock Neural {{Sign Language Translation}}.
\newblock In {\em Proceedings of the {{IEEE Conference}} on {{Computer Vision}}
  and {{Pattern Recognition}}}, pages 7784--7793, 2018.

\bibitem{camgozMultichannelTransformersMultiarticulatory2020}
Necati~Cihan Camgoz, Oscar Koller, Simon Hadfield, and Richard Bowden.
\newblock Multi-channel {{Transformers}} for {{Multi-articulatory Sign Language
  Translation}}.
\newblock In Adrien Bartoli and Andrea Fusiello, editors, {\em Computer
  {{Vision}} \textendash{} {{ECCV}} 2020 {{Workshops}}}, Lecture {{Notes}} in
  {{Computer Science}}, pages 301--319, {Cham}, 2020. {Springer International
  Publishing}.

\bibitem{camgozSignLanguageTransformers2020}
Necati~Cihan Camgoz, Oscar Koller, Simon Hadfield, and Richard Bowden.
\newblock Sign {{Language Transformers}}: {{Joint End-to-End Sign Language
  Recognition}} and {{Translation}}.
\newblock In {\em Proceedings of the {{IEEE}}/{{CVF Conference}} on {{Computer
  Vision}} and {{Pattern Recognition}}}, pages 10023--10033, 2020.

\bibitem{chenMicrosoftCOCOCaptions2015}
Xinlei Chen, Hao Fang, Tsung-Yi Lin, Ramakrishna Vedantam, Saurabh Gupta, Piotr
  Doll{\'a}r, and C.~Lawrence Zitnick.
\newblock Microsoft {{COCO}} captions: {{Data}} collection and evaluation
  server.
\newblock {\em CoRR}, abs/1504.00325, 2015.

\bibitem{chenSimpleMultiModalityTransfer2022}
Yutong Chen, Fangyun Wei, Xiao Sun, Zhirong Wu, and Stephen Lin.
\newblock A {{Simple Multi-Modality Transfer Learning Baseline}} for {{Sign
  Language Translation}}.
\newblock In {\em Proceedings of the {{IEEE}}/{{CVF Conference}} on {{Computer
  Vision}} and {{Pattern Recognition}}}, pages 5120--5130, 2022.

\bibitem{chentwo}
Yutong Chen, Ronglai Zuo, Fangyun Wei, Yu Wu, LIU Shujie, and Brian Mak.
\newblock Two-stream network for sign language recognition and translation.
\newblock In {\em Advances in Neural Information Processing Systems}.

\bibitem{chengFullyConvolutionalNetworks2020}
Ka~Leong Cheng, Zhaoyang Yang, Qifeng Chen, and Yu-Wing Tai.
\newblock Fully {{Convolutional Networks}} for {{Continuous Sign Language
  Recognition}}.
\newblock In Andrea Vedaldi, Horst Bischof, Thomas Brox, and Jan-Michael Frahm,
  editors, {\em Computer {{Vision}} \textendash{} {{ECCV}} 2020}, Lecture
  {{Notes}} in {{Computer Science}}, pages 697--714, {Cham}, 2020. {Springer
  International Publishing}.

\bibitem{choLearningPhraseRepresentations2014}
Kyunghyun Cho, Bart {van Merri{\"e}nboer}, Caglar Gulcehre, Dzmitry Bahdanau,
  Fethi Bougares, Holger Schwenk, and Yoshua Bengio.
\newblock Learning {{Phrase Representations}} using {{RNN
  Encoder}}\textendash{{Decoder}} for {{Statistical Machine Translation}}.
\newblock In {\em Proceedings of the 2014 {{Conference}} on {{Empirical
  Methods}} in {{Natural Language Processing}} ({{EMNLP}})}, pages 1724--1734,
  {Doha, Qatar}, 2014. {Association for Computational Linguistics}.

\bibitem{cihancamgozSubUNetsEndToEndHand2017}
Necati Cihan~Camgoz, Simon Hadfield, Oscar Koller, and Richard Bowden.
\newblock {{SubUNets}}: {{End-To-End Hand Shape}} and {{Continuous Sign
  Language Recognition}}.
\newblock In {\em Proceedings of the {{IEEE International Conference}} on
  {{Computer Vision}}}, pages 3056--3065, 2017.

\bibitem{conneauSupervisedLearningUniversal2017}
Alexis Conneau, Douwe Kiela, Holger Schwenk, Lo{\"i}c Barrault, and Antoine
  Bordes.
\newblock Supervised {{Learning}} of {{Universal Sentence Representations}}
  from {{Natural Language Inference Data}}.
\newblock In {\em Proceedings of the 2017 {{Conference}} on {{Empirical
  Methods}} in {{Natural Language Processing}}}, pages 670--680, {Copenhagen,
  Denmark}, 2017. {Association for Computational Linguistics}.

\bibitem{cuiRecurrentConvolutionalNeural2017}
Runpeng Cui, Hu Liu, and Changshui Zhang.
\newblock Recurrent {{Convolutional Neural Networks}} for {{Continuous Sign
  Language Recognition}} by {{Staged Optimization}}.
\newblock In {\em Proceedings of the {{IEEE Conference}} on {{Computer Vision}}
  and {{Pattern Recognition}}}, pages 7361--7369, 2017.

\bibitem{cuiDeepNeuralFramework2019}
Runpeng Cui, Hu Liu, and Changshui Zhang.
\newblock A {{Deep Neural Framework}} for {{Continuous Sign Language
  Recognition}} by {{Iterative Training}}.
\newblock {\em IEEE Transactions on Multimedia}, 21(7):1880--1891, July 2019.

\bibitem{devlinBERTPretrainingDeep2019}
Jacob Devlin, Ming-Wei Chang, Kenton Lee, and Kristina Toutanova.
\newblock {{BERT}}: {{Pre-training}} of {{Deep Bidirectional Transformers}} for
  {{Language Understanding}}.
\newblock In {\em Proceedings of the 2019 {{Conference}} of the {{North
  American Chapter}} of the {{Association}} for {{Computational Linguistics}}:
  {{Human Language Technologies}}, {{Volume}} 1 ({{Long}} and {{Short
  Papers}})}, pages 4171--4186, {Minneapolis, Minnesota}, 2019. {Association
  for Computational Linguistics}.

\bibitem{ganSkeletonAwareNeuralSign2021}
Shiwei Gan, Yafeng Yin, Zhiwei Jiang, Lei Xie, and Sanglu Lu.
\newblock Skeleton-{{Aware Neural Sign Language Translation}}.
\newblock In {\em Proceedings of the 29th {{ACM International Conference}} on
  {{Multimedia}}}, pages 4353--4361, {New York, NY, USA}, Oct. 2021.
  {Association for Computing Machinery}.

\bibitem{glorotUnderstandingDifficultyTraining2010}
Xavier Glorot and Yoshua Bengio.
\newblock Understanding the difficulty of training deep feedforward neural
  networks.
\newblock In {\em Proceedings of the {{Thirteenth International Conference}} on
  {{Artificial Intelligence}} and {{Statistics}}}, pages 249--256. {JMLR
  Workshop and Conference Proceedings}, Mar. 2010.

\bibitem{glorotDeepSparseRectifier2011}
Xavier Glorot, Antoine Bordes, and Yoshua Bengio.
\newblock Deep {{Sparse Rectifier Neural Networks}}.
\newblock In {\em Proceedings of the {{Fourteenth International Conference}} on
  {{Artificial Intelligence}} and {{Statistics}}}, pages 315--323. {JMLR
  Workshop and Conference Proceedings}, June 2011.

\bibitem{li2023date}
Li Haoyuan, Jiang Hao, Jin Tao, Li Mengyan, Chen Yan, Lin Zhijie, Zhao Yang,
  and Zhao Zhou.
\newblock Date: Domain adaptive product seeker for e-commerce.
\newblock In {\em CVPR}, 2023.

\bibitem{heDeepResidualLearning2016}
Kaiming He, Xiangyu Zhang, Shaoqing Ren, and Jian Sun.
\newblock Deep {{Residual Learning}} for {{Image Recognition}}.
\newblock In {\em Proceedings of the {{IEEE Conference}} on {{Computer Vision}}
  and {{Pattern Recognition}}}, pages 770--778, 2016.

\bibitem{DBLP:conf/lrec/Heinzerling018}
Benjamin Heinzerling and Michael Strube.
\newblock {{BPEmb}}: {{Tokenization-free}} pre-trained subword embeddings in
  275 languages.
\newblock In Nicoletta Calzolari, Khalid Choukri, Christopher Cieri, Thierry
  Declerck, Sara Goggi, K{\^o}iti Hasida, Hitoshi Isahara, Bente Maegaard,
  Joseph Mariani, H{\'e}l{\`e}ne Mazo, Asunci{\'o}n Moreno, Jan Odijk, Stelios
  Piperidis, and Takenobu Tokunaga, editors, {\em Proceedings of the Eleventh
  International Conference on Language Resources and Evaluation, {{LREC}} 2018,
  Miyazaki, Japan, May 7-12, 2018}. {European Language Resources Association
  (ELRA)}, 2018.

\bibitem{huangSignLanguageRecognition2015}
Jie Huang, Wengang Zhou, Houqiang Li, and Weiping Li.
\newblock Sign {{Language Recognition}} using {{3D}} convolutional neural
  networks.
\newblock In {\em 2015 {{IEEE International Conference}} on {{Multimedia}} and
  {{Expo}} ({{ICME}})}, pages 1--6, June 2015.

\bibitem{huangVideoBasedSignLanguage2018}
Jie Huang, Wengang Zhou, Qilin Zhang, Houqiang Li, and Weiping Li.
\newblock Video-{{Based Sign Language Recognition Without Temporal
  Segmentation}}.
\newblock {\em Proceedings of the AAAI Conference on Artificial Intelligence},
  32(1), Apr. 2018.

\bibitem{huanggenerspeech}
Rongjie Huang, Yi Ren, Jinglin Liu, Chenye Cui, and Zhou Zhao.
\newblock Generspeech: Towards style transfer for generalizable out-of-domain
  text-to-speech.
\newblock In {\em Advances in Neural Information Processing Systems}.

\bibitem{huang2022prodiff}
Rongjie Huang, Zhou Zhao, Huadai Liu, Jinglin Liu, Chenye Cui, and Yi Ren.
\newblock Prodiff: Progressive fast diffusion model for high-quality
  text-to-speech.
\newblock In {\em Proceedings of the 30th ACM International Conference on
  Multimedia}, pages 2595--2605, 2022.

\bibitem{huang2022transpeech}
Rongjie Huang, Zhou Zhao, Jinglin Liu, Huadai Liu, Yi Ren, Lichao Zhang, and
  Jinzheng He.
\newblock Transpeech: Speech-to-speech translation with bilateral perturbation.
\newblock {\em arXiv preprint arXiv:2205.12523}, 2022.

\bibitem{ioffeBatchNormalizationAccelerating2015}
Sergey Ioffe and Christian Szegedy.
\newblock Batch {{Normalization}}: {{Accelerating Deep Network Training}} by
  {{Reducing Internal Covariate Shift}}.
\newblock In {\em Proceedings of the 32nd {{International Conference}} on
  {{Machine Learning}}}, pages 448--456. {PMLR}, June 2015.

\bibitem{jin2021contrastive}
Tao Jin and Zhou Zhao.
\newblock Contrastive disentangled meta-learning for signer-independent sign
  language translation.
\newblock In {\em Proceedings of the 29th ACM International Conference on
  Multimedia}, pages 5065--5073, 2021.

\bibitem{jin2022mc}
Tao Jin, Zhou Zhao, Meng Zhang, and Xingshan Zeng.
\newblock Mc-slt: Towards low-resource signer-adaptive sign language
  translation.
\newblock In {\em Proceedings of the 30th ACM International Conference on
  Multimedia}, pages 4939--4947, 2022.

\bibitem{jin2022prior}
Tao Jin, Zhou Zhao, Meng Zhang, and Xingshan Zeng.
\newblock Prior knowledge and memory enriched transformer for sign language
  translation.
\newblock In {\em Findings of the Association for Computational Linguistics:
  ACL 2022}, pages 3766--3775, 2022.

\bibitem{kingmaAdamMethodStochastic2015}
Diederik~P. Kingma and Jimmy Ba.
\newblock Adam: {{A}} method for stochastic optimization.
\newblock In Yoshua Bengio and Yann LeCun, editors, {\em 3rd International
  Conference on Learning Representations, {{ICLR}} 2015, San Diego, {{CA}},
  {{USA}}, May 7-9, 2015, Conference Track Proceedings}, 2015.

\bibitem{kirosSkipThoughtVectors2015}
Ryan Kiros, Yukun Zhu, Russ~R Salakhutdinov, Richard Zemel, Raquel Urtasun,
  Antonio Torralba, and Sanja Fidler.
\newblock Skip-{{Thought Vectors}}.
\newblock In {\em Advances in {{Neural Information Processing Systems}}},
  volume~28. {Curran Associates, Inc.}, 2015.

\bibitem{koNeuralSignLanguage2019}
Sang-Ki Ko, Chang~Jo Kim, Hyedong Jung, and Choongsang Cho.
\newblock Neural {{Sign Language Translation Based}} on {{Human Keypoint
  Estimation}}.
\newblock {\em Applied Sciences}, 9(13):2683, Jan. 2019.

\bibitem{koller2019weakly}
Oscar Koller, Necati~Cihan Camgoz, Hermann Ney, and Richard Bowden.
\newblock Weakly supervised learning with multi-stream cnn-lstm-hmms to
  discover sequential parallelism in sign language videos.
\newblock {\em IEEE transactions on pattern analysis and machine intelligence},
  42(9):2306--2320, 2019.

\bibitem{kreutzer-etal-2019-joey}
Julia Kreutzer, Jasmijn Bastings, and Stefan Riezler.
\newblock Joey {NMT}: A minimalist {NMT} toolkit for novices.
\newblock In {\em Proceedings of the 2019 Conference on Empirical Methods in
  Natural Language Processing and the 9th International Joint Conference on
  Natural Language Processing (EMNLP-IJCNLP): System Demonstrations}, pages
  109--114, Hong Kong, China, Nov. 2019. Association for Computational
  Linguistics.

\bibitem{kudoSentencePieceSimpleLanguage2018}
Taku Kudo and John Richardson.
\newblock {{SentencePiece}}: {{A}} simple and language independent subword
  tokenizer and detokenizer for {{Neural Text Processing}}.
\newblock In {\em Proceedings of the 2018 {{Conference}} on {{Empirical
  Methods}} in {{Natural Language Processing}}: {{System Demonstrations}}},
  pages 66--71, {Brussels, Belgium}, 2018. {Association for Computational
  Linguistics}.

\bibitem{leDistributedRepresentationsSentences2014}
Quoc Le and Tomas Mikolov.
\newblock Distributed {{Representations}} of {{Sentences}} and {{Documents}}.
\newblock In {\em Proceedings of the 31st {{International Conference}} on
  {{Machine Learning}}}, pages 1188--1196. {PMLR}, June 2014.

\bibitem{liWordlevelDeepSign2020}
Dongxu Li, Cristian Rodriguez, Xin Yu, and Hongdong Li.
\newblock Word-level {{Deep Sign Language Recognition}} from {{Video}}: {{A New
  Large-scale Dataset}} and {{Methods Comparison}}.
\newblock In {\em Proceedings of the {{IEEE}}/{{CVF Winter Conference}} on
  {{Applications}} of {{Computer Vision}}}, pages 1459--1469, 2020.

\bibitem{liTransferringCrossDomainKnowledge2020}
Dongxu Li, Xin Yu, Chenchen Xu, Lars Petersson, and Hongdong Li.
\newblock Transferring {{Cross-Domain Knowledge}} for {{Video Sign Language
  Recognition}}.
\newblock In {\em Proceedings of the {{IEEE}}/{{CVF Conference}} on {{Computer
  Vision}} and {{Pattern Recognition}}}, pages 6205--6214, 2020.

\bibitem{liTSPNetHierarchicalFeature2020}
{\relax DONGXU} LI, Chenchen Xu, Xin Yu, Kaihao Zhang, Benjamin Swift, Hanna
  Suominen, and Hongdong Li.
\newblock {{TSPNet}}: {{Hierarchical Feature Learning}} via {{Temporal Semantic
  Pyramid}} for {{Sign Language Translation}}.
\newblock In {\em Advances in {{Neural Information Processing Systems}}},
  volume~33, pages 12034--12045. {Curran Associates, Inc.}, 2020.

\bibitem{linAutomaticEvaluationMachine2004}
Chin-Yew Lin and Franz~Josef Och.
\newblock Automatic {{Evaluation}} of {{Machine Translation Quality Using
  Longest Common Subsequence}} and {{Skip-Bigram Statistics}}.
\newblock In {\em Proceedings of the 42nd {{Annual Meeting}} of the
  {{Association}} for {{Computational Linguistics}} ({{ACL-04}})}, pages
  605--612, {Barcelona, Spain}, 2004.

\bibitem{lin2021simullr}
Zhijie Lin, Zhou Zhao, Haoyuan Li, Jinglin Liu, Meng Zhang, Xingshan Zeng, and
  Xiaofei He.
\newblock Simullr: Simultaneous lip reading transducer with attention-guided
  adaptive memory.
\newblock In {\em Proceedings of the 29th ACM International Conference on
  Multimedia}, pages 1359--1367, 2021.

\bibitem{luongEffectiveApproachesAttentionbased2015}
Thang Luong, Hieu Pham, and Christopher~D. Manning.
\newblock Effective {{Approaches}} to {{Attention-based Neural Machine
  Translation}}.
\newblock In {\em Proceedings of the 2015 {{Conference}} on {{Empirical
  Methods}} in {{Natural Language Processing}}}, pages 1412--1421, {Lisbon,
  Portugal}, 2015. {Association for Computational Linguistics}.

\bibitem{maSTACLSimultaneousTranslation2019}
Mingbo Ma, Liang Huang, Hao Xiong, Renjie Zheng, Kaibo Liu, Baigong Zheng,
  Chuanqiang Zhang, Zhongjun He, Hairong Liu, Xing Li, Hua Wu, and Haifeng
  Wang.
\newblock {{STACL}}: {{Simultaneous Translation}} with {{Implicit
  Anticipation}} and {{Controllable Latency}} using {{Prefix-to-Prefix
  Framework}}.
\newblock In {\em Proceedings of the 57th {{Annual Meeting}} of the
  {{Association}} for {{Computational Linguistics}}}, pages 3025--3036,
  {Florence, Italy}, 2019. {Association for Computational Linguistics}.

\bibitem{martinezPurdueRVLSLLLASL2002}
A.M. Martinez, R.B. Wilbur, R. Shay, and A.C. Kak.
\newblock Purdue {{RVL-SLLL ASL}} database for automatic recognition of
  {{American Sign Language}}.
\newblock In {\em Proceedings. {{Fourth IEEE International Conference}} on
  {{Multimodal Interfaces}}}, pages 167--172, 2002.

\bibitem{minVisualAlignmentConstraint2021a}
Yuecong Min, Aiming Hao, Xiujuan Chai, and Xilin Chen.
\newblock Visual {{Alignment Constraint}} for {{Continuous Sign Language
  Recognition}}.
\newblock In {\em Proceedings of the {{IEEE}}/{{CVF International Conference}}
  on {{Computer Vision}}}, pages 11542--11551, 2021.

\bibitem{mitchell1980need}
Tom~M Mitchell.
\newblock {\em The need for biases in learning generalizations}.
\newblock Department of Computer Science, Laboratory for Computer Science
  Research~…, 1980.

\bibitem{ong2005automatic}
Sylvie~CW Ong and Surendra Ranganath.
\newblock Automatic sign language analysis: {{A}} survey and the future beyond
  lexical meaning.
\newblock {\em IEEE Transactions on Pattern Analysis \& Machine Intelligence},
  27(06):873--891, 2005.

\bibitem{orbayNeuralSignLanguage2020}
Alptekin Orbay and Lale Akarun.
\newblock Neural {{Sign Language Translation}} by {{Learning Tokenization}}.
\newblock In {\em 2020 15th {{IEEE International Conference}} on {{Automatic
  Face}} and {{Gesture Recognition}} ({{FG}} 2020)}, pages 222--228, 2020.

\bibitem{web1}
World~Health Organization.
\newblock Deafness and hearing loss.
\newblock
  \url{https://www.who.int/news-room/fact-sheets/detail/deafness-and-hearing-loss},
  2021.

\bibitem{papineniBleuMethodAutomatic2002}
Kishore Papineni, Salim Roukos, Todd Ward, and Wei-Jing Zhu.
\newblock Bleu: A {{Method}} for {{Automatic Evaluation}} of {{Machine
  Translation}}.
\newblock In {\em Proceedings of the 40th {{Annual Meeting}} of the
  {{Association}} for {{Computational Linguistics}}}, pages 311--318,
  {Philadelphia, Pennsylvania, USA}, 2002. {Association for Computational
  Linguistics}.

\bibitem{paszkePyTorchImperativeStyle2019}
Adam Paszke, Sam Gross, Francisco Massa, Adam Lerer, James Bradbury, Gregory
  Chanan, Trevor Killeen, Zeming Lin, Natalia Gimelshein, Luca Antiga, Alban
  Desmaison, Andreas Kopf, Edward Yang, Zachary DeVito, Martin Raison, Alykhan
  Tejani, Sasank Chilamkurthy, Benoit Steiner, Lu Fang, Junjie Bai, and Soumith
  Chintala.
\newblock {{PyTorch}}: {{An Imperative Style}}, {{High-Performance Deep
  Learning Library}}.
\newblock In {\em Advances in {{Neural Information Processing Systems}}},
  volume~32. {Curran Associates, Inc.}, 2019.

\bibitem{pfau2018syntax}
Roland Pfau, Martin Salzmann, and Markus Steinbach.
\newblock The syntax of sign language agreement: Common ingredients, but
  unusual recipe.
\newblock {\em Glossa: a journal of general linguistics}, 3(1), 2018.

\bibitem{reimersSentenceBERTSentenceEmbeddings2019}
Nils Reimers and Iryna Gurevych.
\newblock Sentence-{{BERT}}: {{Sentence Embeddings}} using {{Siamese
  BERT-Networks}}.
\newblock In {\em Proceedings of the 2019 {{Conference}} on {{Empirical
  Methods}} in {{Natural Language Processing}} and the 9th {{International
  Joint Conference}} on {{Natural Language Processing}} ({{EMNLP-IJCNLP}})},
  pages 3982--3992, {Hong Kong, China}, 2019. {Association for Computational
  Linguistics}.

\bibitem{ruckleConcatenatedPowerMean2018}
Andreas R{\"u}ckl{\'e}, Steffen Eger, Maxime Peyrard, and Iryna Gurevych.
\newblock Concatenated {{Power Mean Word Embeddings}} as {{Universal
  Cross-Lingual Sentence Representations}}, Sept. 2018.

\bibitem{sennrichNeuralMachineTranslation2016}
Rico Sennrich, Barry Haddow, and Alexandra Birch.
\newblock Neural {{Machine Translation}} of {{Rare Words}} with {{Subword
  Units}}.
\newblock In {\em Proceedings of the 54th {{Annual Meeting}} of the
  {{Association}} for {{Computational Linguistics}} ({{Volume}} 1: {{Long
  Papers}})}, pages 1715--1725, {Berlin, Germany}, 2016. {Association for
  Computational Linguistics}.

\bibitem{shimUnderstandingRoleSelf2021}
Kyuhong Shim, Jungwook Choi, and Wonyong Sung.
\newblock Understanding the {{Role}} of {{Self Attention}} for {{Efficient
  Speech Recognition}}.
\newblock In {\em International {{Conference}} on {{Learning
  Representations}}}, Sept. 2021.

\bibitem{starnerRealtimeAmericanSign1998}
T. Starner, J. Weaver, and A. Pentland.
\newblock Real-time {{American}} sign language recognition using desk and
  wearable computer based video.
\newblock {\em IEEE Transactions on Pattern Analysis and Machine Intelligence},
  20(12):1371--1375, 1998.

\bibitem{szegedyRethinkingInceptionArchitecture2016}
Christian Szegedy, Vincent Vanhoucke, Sergey Ioffe, Jon Shlens, and Zbigniew
  Wojna.
\newblock Rethinking the {{Inception Architecture}} for {{Computer Vision}}.
\newblock In {\em Proceedings of the {{IEEE Conference}} on {{Computer Vision}}
  and {{Pattern Recognition}}}, pages 2818--2826, 2016.

\bibitem{vaezijoze2019ms-asl}
Hamid Vaezi~Joze and Oscar Koller.
\newblock {{MS-ASL}}: {{A}} large-scale data set and benchmark for
  understanding american sign language.
\newblock In {\em The British Machine Vision Conference ({{BMVC}})}, Sept.
  2019.

\bibitem{vaswaniAttentionAllYou2017}
Ashish Vaswani, Noam Shazeer, Niki Parmar, Jakob Uszkoreit, Llion Jones,
  Aidan~N Gomez, {\L}ukasz Kaiser, and Illia Polosukhin.
\newblock Attention is {{All}} you {{Need}}.
\newblock In {\em Advances in {{Neural Information Processing Systems}}},
  volume~30. {Curran Associates, Inc.}, 2017.

\bibitem{voskou2021stochastic}
Andreas Voskou, Konstantinos~P Panousis, Dimitrios Kosmopoulos, Dimitris~N
  Metaxas, and Sotirios Chatzis.
\newblock Stochastic transformer networks with linear competing units:
  Application to end-to-end sl translation.
\newblock In {\em Proceedings of the IEEE/CVF International Conference on
  Computer Vision}, pages 11946--11955, 2021.

\bibitem{xia2022video}
Yan Xia, Zhou Zhao, Shangwei Ye, Yang Zhao, Haoyuan Li, and Yi Ren.
\newblock Video-guided curriculum learning for spoken video grounding.
\newblock In {\em Proceedings of the 30th ACM International Conference on
  Multimedia}, pages 5191--5200, 2022.

\bibitem{yinMLSLTMultilingualSign2022b}
Aoxiong Yin, Zhou Zhao, Weike Jin, Meng Zhang, Xingshan Zeng, and Xiaofei He.
\newblock {{MLSLT}}: {{Towards}} multilingual sign language translation.
\newblock In {\em {{IEEE}}/{{CVF}} Conference on Computer Vision and Pattern
  Recognition, {{CVPR}} 2022, New Orleans, {{LA}}, {{USA}}, June 18-24, 2022},
  pages 5099--5109. {IEEE}, 2022.

\bibitem{yinSimulSLTEndtoEndSimultaneous2021}
Aoxiong Yin, Zhou Zhao, Jinglin Liu, Weike Jin, Meng Zhang, Xingshan Zeng, and
  Xiaofei He.
\newblock {{SimulSLT}}: {{End-to-End Simultaneous Sign Language Translation}}.
\newblock In {\em Proceedings of the 29th {{ACM International Conference}} on
  {{Multimedia}}}, pages 4118--4127, {New York, NY, USA}, Oct. 2021.
  {Association for Computing Machinery}.

\bibitem{yinBetterSignLanguage2020}
Kayo Yin and Jesse Read.
\newblock Better {{Sign Language Translation}} with {{STMC-Transformer}}.
\newblock In {\em Proceedings of the 28th {{International Conference}} on
  {{Computational Linguistics}}}, pages 5975--5989, {Barcelona, Spain
  (Online)}, 2020. {International Committee on Computational Linguistics}.

\bibitem{zaheer2020big}
Manzil Zaheer, Guru Guruganesh, Kumar~Avinava Dubey, Joshua Ainslie, Chris
  Alberti, Santiago Ontanon, Philip Pham, Anirudh Ravula, Qifan Wang, Li Yang,
  et~al.
\newblock Big bird: Transformers for longer sequences.
\newblock {\em Advances in neural information processing systems},
  33:17283--17297, 2020.

\bibitem{zhouImprovingSignLanguage2021}
Hao Zhou, Wengang Zhou, Weizhen Qi, Junfu Pu, and Houqiang Li.
\newblock Improving {{Sign Language Translation With Monolingual Data}} by
  {{Sign Back-Translation}}.
\newblock In {\em Proceedings of the {{IEEE}}/{{CVF Conference}} on {{Computer
  Vision}} and {{Pattern Recognition}}}, pages 1316--1325, 2021.

\bibitem{zhouSpatialTemporalMultiCueNetwork2020}
Hao Zhou, Wengang Zhou, Yun Zhou, and Houqiang Li.
\newblock Spatial-{{Temporal Multi-Cue Network}} for {{Continuous Sign Language
  Recognition}}.
\newblock {\em Proceedings of the AAAI Conference on Artificial Intelligence},
  34(07):13009--13016, Apr. 2020.

\end{thebibliography}
}

\end{document}